\documentclass[letterpaper]{article} 
\usepackage{iclr2027_conference,times}

\usepackage{amsmath,amssymb}
\usepackage{graphicx}
\usepackage[table]{xcolor}
\usepackage{booktabs,multirow,makecell}
\usepackage{adjustbox}
\usepackage{pgfplots}
\pgfplotsset{compat=1.18}

\usepackage{amsmath,amsfonts,bm}

\def\eqref#1{equation~\ref{#1}}

\def\1{\bm{1}}

\DeclareMathAlphabet{\mathsfit}{\encodingdefault}{\sfdefault}{m}{sl}
\SetMathAlphabet{\mathsfit}{bold}{\encodingdefault}{\sfdefault}{bx}{n}

\providecommand{\eg}{\emph{e.g.}}

\usepackage{hyperref}
\usepackage{url}
\usepackage{enumitem}
\usepackage{cleveref}
\usepackage{algorithm}
\usepackage{algorithmic}
\usepackage{caption}
\title{COMPASS: Contextual Moment Prior Alignment for Generalizable AIGI Detection}

\author{
\normalfont\large
Zhan Yang$^{1*}$ \quad Tiancheng Shen$^{2*}$ \quad Weijie Lyu$^{2}$ \quad Shuo Chen$^{2}$ \\[3pt]
\normalfont\large
Bernard Ghanem$^{3}$ \quad Yibo Yang$^{4}$ \quad Ming-Hsuan Yang$^{2}$ \\[10pt]
\normalfont\normalsize
$^{1}$Zhejiang University \quad $^{2}$University of California, Merced \\[2pt]
\normalfont\normalsize
$^{3}$KAUST \quad $^{4}$Shanghai Jiao Tong University \\[8pt]
\normalfont\small
$^{*}$Equal contribution.
}

\iclrfinalcopy 
\begin{document}

\maketitle

\begin{abstract}
Generalizable AI-generated image (AIGI) and deepfake detection remains challenging due to the \emph{asymmetry phenomenon}, where adaptation to limited training distributions can degrade the transferable representations of Vision Foundation Models (VFMs). Recent subspace adaptation methods mitigate this issue by freezing high-magnitude principal components while adapting the residual subspace. However, such magnitude-based decomposition is data-agnostic and does not explicitly account for target-relevant semantics, potentially leading to \emph{semantic misalignment}. We propose \textbf{COMPASS} (COntextual Moment Prior Alignment for Subspace Separation), a data-aware subspace adaptation framework that uses a small unlabeled \emph{Semantic Proxy Set} to estimate target activation statistics. COMPASS incorporates activation covariance into the decomposition of pre-trained weights, preserving target-relevant representations as a stable semantic reference while the residual adapter learns forgery-related variations. The proxy calibration requires no gradient-based optimization. Extensive experiments demonstrate strong generalization across cross-dataset, cross-method, joint cross-dataset-method, and open-world cross-generator evaluations, achieving state-of-the-art performance on both Deepfake and AIGI detection benchmarks. Source code and trained models will be publicly released.
\end{abstract}

\section{Introduction}
\label{sec:intro}

Recent advances in generative AI~\citep{goodfellow2020generative,ddpm,rombach2022high,yan2025gpt,chen2026context,shen2025qk} enable hyper-realistic synthesis, posing security risks such as deepfakes and visual disinformation~\citep{korshunov2018deepfakes,thies2016face2face,shoaib2023deepfakes,singh2025unmasking,easttom2025deepfake}. 
Consequently, robust AI-generated image (AIGI) detection is critical~\citep{wang2020cnn,rossler2019faceforensics++,yan2025effort,sarkar2025unsupervised,rathore2021image,fridrich2003detection,kashyap2017detection}. 
A key challenge is the \emph{asymmetry phenomenon}: detectors overfit to limited forgery patterns and biases~\citep{yan2025effort}, yielding entangled features that can hinder generalization to unseen manipulations.

To address this issue, recent works explore low-rank fine-tuning of Vision Foundation Models (VFMs)~\citep{clip} via Orthogonal Subspace Decomposition~\citep{wall2003singular}. 
EFFORT~\citep{yan2025effort}, a state-of-the-art SVD-based AIGI detector, emphasizes preserving pre-trained VFM knowledge for generalization. 
It applies SVD to decompose the pre-trained weights into principal and residual components, freezing principal components with the largest magnitudes to retain core semantics, while fine-tuning residual components to capture forgery-specific patterns.

The motivation for preserving pre-trained semantics is not to use semantic content itself as a cue for real/fake classification. Rather, real and generated images typically share the same dominant semantic content, while forgery-specific signals appear as variations within this shared context. Preserving the principal semantic subspace therefore provides a stable reference for the residual adapter to learn forgery-related variations without unnecessarily altering the underlying content representation.

However, its magnitude-based implementation has an important limitation: it prioritizes components according to their contribution to global weight reconstruction. Although mathematically well-founded, this criterion is data-agnostic and does not explicitly account for task-specific semantics.
In AIGI detection, substantial distribution shifts arise across dataset domains (e.g., faces or human bodies), manipulation types, and generative models. 
As a result, preserving high-magnitude components does not guarantee retention of information most relevant for robust forgery detection.

\begin{figure}[tp]
  \centering\includegraphics[width=\linewidth]{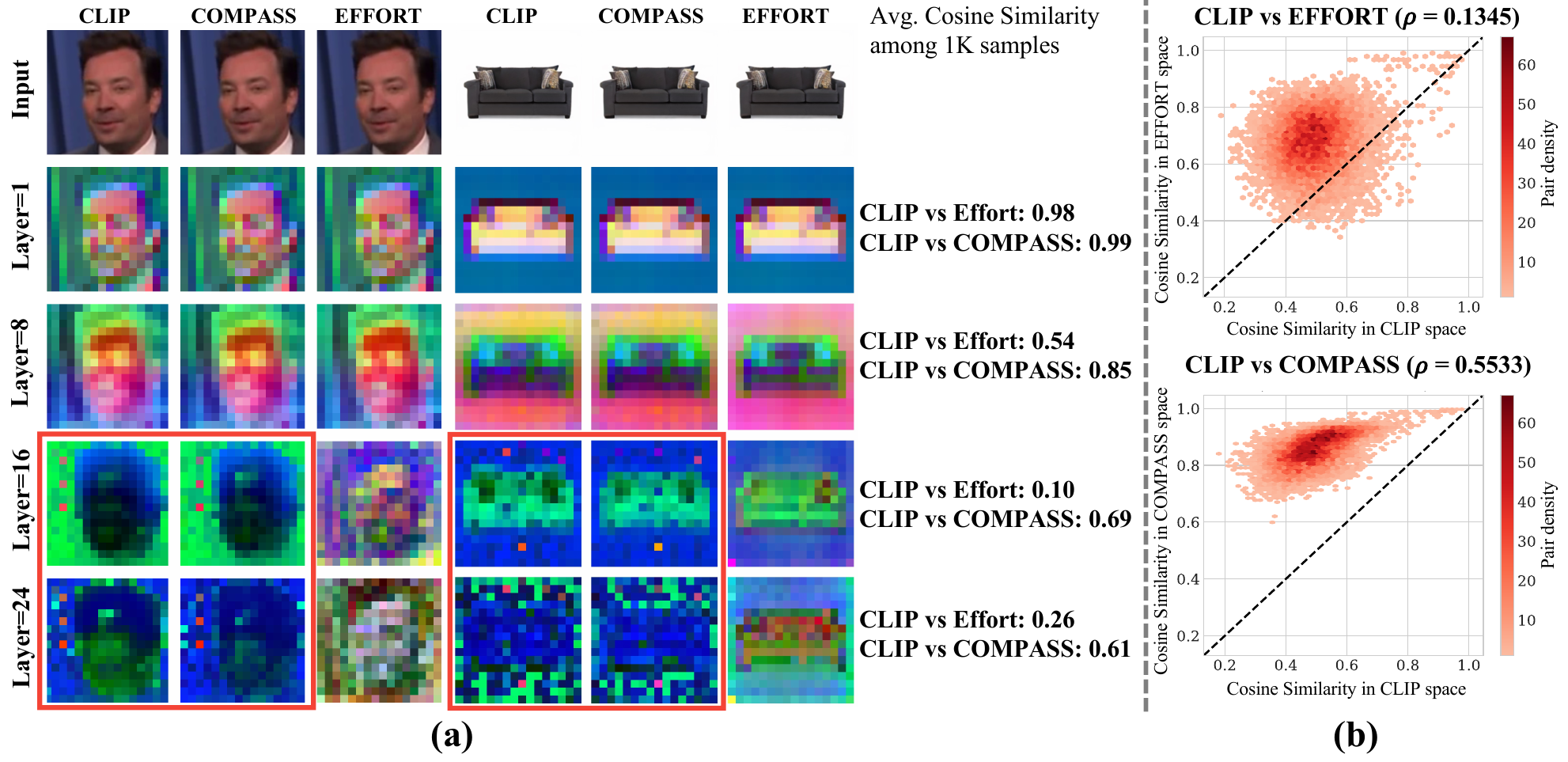}
  \caption{Pilot study on feature retention prior to fine-tuning.
(a) PCA visualizations of intermediate features across 1K images. Magnitude-based truncation in EFFORT increasingly alters the original feature geometry in deeper layers, with the average cosine similarity to the original CLIP representation dropping to 0.26 in Layer 24, compared with 0.61 for COMPASS.
(b) Pairwise similarity density plots using only the principal subspace ($\mathbf{W}_{prin}$, before fine-tuning). Each point compares the cosine similarity of the same \textbf{image pair} in the original CLIP and principal-only spaces. COMPASS better preserves the relative similarity structure, achieving a higher Spearman rank correlation ($\rho_s=0.5533$) than EFFORT ($\rho_s=0.1345$). Since truncation and feature re-normalization may shift absolute cosine similarities, we focus on rank preservation rather than proximity to the diagonal; further analysis is provided in \cref{sec:math_offset}.}
  \label{fig:pilot}
  \vspace{-2mm}
\end{figure}

Our pilot analysis in \cref{fig:pilot} illustrates this limitation. Using feature retention, a metric adopted by prior orthogonal adaptation methods~\citep{yan2025effort}, we find that magnitude-based decomposition can substantially alter the pre-trained representation retained by the principal subspace. We refer to this mismatch between magnitude-based preservation and target-relevant information as \emph{semantic misalignment}. The results further suggest that incorporating target-distribution activation statistics can mitigate this mismatch and improve the preservation of pre-trained representations.

Building on this observation, we introduce \textbf{COMPASS} (COntextual Moment Prior Alignment for Subspace Separation), a data-aware subspace adaptation framework that incorporates proxy activation statistics into subspace partitioning. Before fine-tuning, we sample a small unlabeled \emph{Semantic Proxy Set} (e.g., $\sim$80 images) from the target-category distribution and compute its activation covariance to guide the SVD of pre-trained weights. This lightweight, zero-backpropagation calibration biases the preserved subspace toward directions emphasized by the proxy activation distribution. The proxy samples are used only to estimate activation statistics for subspace calibration and do not participate in gradient-based optimization.

Notably, this proxy-guided calibration better preserves pre-trained representations than magnitude-based decomposition. Even when the Semantic Proxy Set is constructed from only one or two datasets within the target category, its activation statistics provide a useful estimate of target-relevant activation structure that remains effective across different data sources. By retaining these target-relevant pre-trained semantics more faithfully, COMPASS establishes a more stable reference for residual fine-tuning, allowing forgery-related variations to be learned with less interference from source-specific content biases. This improved preservation is consistent with the stronger generalization observed across dataset domains, manipulation types, and generative models.

We evaluate COMPASS under three challenging generalization settings ~\cref{tab:deepfake_results,tab:joint_results,tab:aigi_bench}, where it achieves state-of-the-art performance across (1) \textbf{cross-dataset} scenarios, (2) \textbf{cross-method} scenarios, (3) joint \textbf{cross-dataset-method} scenarios.

The main contributions of this work are:
\begin{itemize}[leftmargin=*, labelindent=0pt]
    \item We identify an important limitation of magnitude-based orthogonal subspace decomposition for AIGI detection: its data-agnostic partitioning may fail to preserve task-relevant pre-trained information. Our pilot analyses show that incorporating proxy-guided calibration can mitigate this \emph{semantic misalignment} and improve feature retention.
    \item We introduce \textbf{COMPASS} (COntextual Moment Prior Alignment for Subspace Separation), which leverages a small unlabeled \emph{Semantic Proxy Set} to compute activation covariances and uses activation covariance to calibrate subspace preservation before fine-tuning. This lightweight, zero-backpropagation design enables data-aware preservation of pre-trained semantic representations.
    \item We demonstrate that COMPASS achieves state-of-the-art generalization across cross-dataset, cross-method, and joint cross-dataset-method evaluations. Our results support the use of a better-preserved semantic reference for residual forgery learning and show consistent improvements across diverse domains, manipulation types, and generative models.
\end{itemize}

\section{Related Work}
\label{sec:related}

\noindent \textbf{Conventional Approaches for AIGI Detection.}
Conventional deepfake and synthetic-image detectors primarily exploit task-specific artifacts, including frequency and spatial cues~\citep{li2021frequency,liu2021spatial,tan2024rethinking}, blending inconsistencies~\citep{shiohara2022detecting,chen2022self}, reconstruction signals~\citep{cao2022end,wang2023dire,luo2024lare}, and content or identity representations~\citep{yan2023ucf,huang2023implicit}. Recent methods further improve artifact-oriented modeling~\citep{liang2025ferretnet,wang2025lota,zhou2025beyond}. Despite strong performance, such task-specific cues may generalize poorly to unseen synthesis processes, motivating increasing use of Vision Foundation Models (VFMs) for more transferable forgery detection.

\noindent \textbf{Vision Foundation Models and Subspace Adaptation.}
Recent methods leverage VFMs such as CLIP~\citep{clip} to improve generalization in forgery detection~\citep{ojha2023towards,wu2023generalizable,liu2024forgery}. However, direct fine-tuning may alter pre-trained representations and reduce cross-domain generalization. EFFORT~\citep{yan2025effort} addresses this issue by freezing high-magnitude SVD components while adapting the residual subspace. Its magnitude-based decomposition, however, remains data-agnostic. COMPASS instead uses target activation statistics to guide subspace preservation.

\section{Method}
\label{sec:method}

\begin{figure}[tb]
  \centering
  \includegraphics[width=\linewidth]{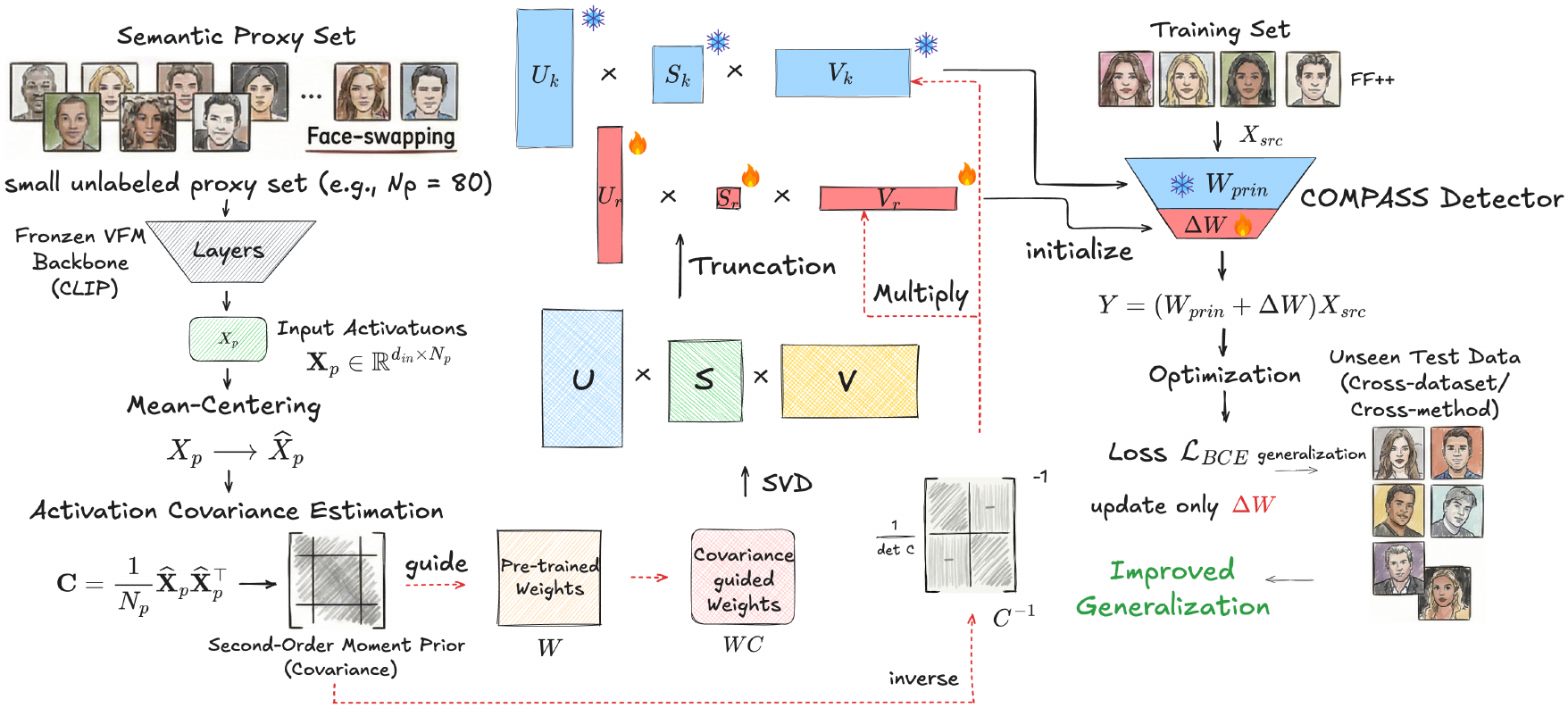}
  \caption{Overall pipeline of COMPASS.
\textbf{Left:} A small unlabeled Semantic Proxy Set is used to estimate the activation covariance $\mathbf{C}$.
\textbf{Middle:} The pre-trained weights are factorized via covariance-guided SVD into principal weights $\mathbf{W}_{prin}$ and residual weights $\mathbf{W}_{res}$.
\textbf{Right:} During fine-tuning, $\mathbf{W}_{prin}$ is frozen as a stable semantic reference, while the residual adapter $\Delta\mathbf{W}$, initialized from $\mathbf{W}_{res}$, is optimized to learn forgery-related variations.}
  \label{fig:framework}
\end{figure}

We present \textbf{COMPASS} (\textbf{CO}ntextual \textbf{M}oment \textbf{P}rior \textbf{A}lignment for \textbf{S}ubspace \textbf{S}eparation), a data-aware subspace adaptation framework for generalizable AIGI detection. As illustrated in \cref{fig:framework}, COMPASS uses a small unlabeled Semantic Proxy Set to estimate target activation statistics, which guide the decomposition of pre-trained weights into principal and residual weight components. The principal weight component is kept fixed as a stable semantic reference, while the residual adapter is optimized to learn forgery-related variations.

\subsection{Preliminaries: Magnitude-Based Subspace Decomposition}
\label{sec:preliminaries}

Given a pre-trained Vision Foundation Model (\eg, CLIP~\citep{clip}) employed as the backbone for deepfake detection, let
$\mathbf{W}\in\mathbb{R}^{d_{out}\times d_{in}}$
denote the weight matrix of a linear layer. EFFORT~\citep{yan2025effort}
decomposes $\mathbf{W}$ using Singular Value Decomposition (SVD):
\begin{equation}
    \mathbf{W}
    =
    \mathbf{U}\mathbf{S}\mathbf{V}^{\top}
    =
    \mathbf{U}_k\mathbf{S}_k\mathbf{V}_k^{\top}
    +
    \mathbf{U}_r\mathbf{S}_r\mathbf{V}_r^{\top},
\end{equation}
where the top-$k$ components are frozen to preserve pre-trained representations, while the residual components are adapted for forgery detection.

This decomposition follows the standard rank-$k$ reconstruction objective
\begin{equation}
    \min_{\operatorname{rank}(\mathbf{W}_k)\leq k}
    \|\mathbf{W}-\mathbf{W}_k\|_F^2.
\end{equation}
Consequently, the retained components are determined solely by their
contribution to reconstructing the weight matrix and are independent of
the input data distribution. In particular, a large singular value does
not necessarily imply that the corresponding direction is important for
preserving the representations activated by the target data.

As shown in \cref{fig:pilot}, magnitude-based SVD can substantially alter the VFM representations retained by the principal subspace, illustrating this \emph{semantic misalignment} and motivating a data-aware response-preservation objective.

\subsection{Proxy-Guided Data-Aware Subspace Calibration}
\label{sec:compass_formulation}

\noindent \textbf{Data-Aware Response Preservation.}
How can the decomposition be made aware of the target data distribution? A direct approach preserves the response of each pre-trained linear layer on target-relevant inputs. Let $\mathbf{x}\in\mathbb{R}^{d_{in}}$ denote the input activation to a layer with original and decomposed weights $\mathbf{W}$ and $\mathbf{W}_k$.
We therefore consider the response-preservation objective
\begin{equation}
\min_{\operatorname{rank}(\mathbf{W}_k)\leq k}
\mathbb{E}_{\mathbf{x}}
\left[
\left\|(\mathbf{W}-\mathbf{W}_k)\mathbf{x}\right\|_2^2
\right].
\end{equation}
This objective is consistent with the motivation of our pilot study in \cref{fig:pilot}, which evaluates whether the preserved principal subspace retains the representations produced by the original VFM.

Assuming that the input activations are mean-centered, let
\begin{equation}
\mathbf{C}
=
\mathbb{E}_{\mathbf{x}}
\left[
\mathbf{x}\mathbf{x}^{\top}
\right]
\end{equation}
denote their covariance matrix. The response-preservation objective can then
be rewritten as
\begin{align}
\mathbb{E}_{\mathbf{x}}
\left[
\left\|(\mathbf{W}-\mathbf{W}_k)\mathbf{x}\right\|_2^2
\right]
&=
\operatorname{Tr}
\left[
(\mathbf{W}-\mathbf{W}_k)
\mathbf{C}
(\mathbf{W}-\mathbf{W}_k)^\top
\right] =
\left\|
(\mathbf{W}-\mathbf{W}_k)\mathbf{C}^{1/2}
\right\|_F^2.
\end{align}
This form motivates performing the low-rank decomposition in the covariance-weighted space \(\mathbf W\mathbf C^{1/2}\), rather than on \(\mathbf W\) alone.
Hence, $\mathbf{C}^{1/2}$ naturally serves as a data-dependent guidance matrix: unlike standard SVD, the decomposition assigns greater importance to directions that are more strongly expressed under the target activation distribution.

\noindent \textbf{Proxy-Based Covariance Estimation.}
To estimate the target activation covariance in practice, we introduce a small unlabeled \emph{Semantic Proxy Set} $\mathcal{D}_{p}$ containing $N_p$ samples drawn from the target-category distribution. Passing these samples through the frozen VFM, we collect the input activations $\mathbf{X}_p \in \mathbb{R}^{d_{in} \times N_p}$ for each target linear layer. After mean-centering the activations to obtain $\widehat{\mathbf{X}}_p$, we estimate the covariance as
\begin{equation}
    \mathbf{C}
    =
    \frac{1}{N_p}
    \widehat{\mathbf{X}}_p
    \widehat{\mathbf{X}}_p^\top.
\end{equation}
We use this covariance as a second-order moment prior that summarizes the activation structure of the proxy distribution. The proxy set is used only to estimate these activation statistics for decomposition and does not participate in gradient-based optimization.

According to the response-preservation objective derived above, $\mathbf{C}^{1/2}$ naturally serves as the \emph{proxy guidance matrix} for making the decomposition data-aware. In practice, however, our experiments show that directly using $\mathbf{C}$ as the guidance matrix provides better generalization than $\mathbf{C}^{1/2}$ (\cref{tab:ablation_generalization}(a)). This choice is empirical rather than directly implied by the preceding derivation. We hypothesize that the stronger covariance weighting of $\mathbf{C}$ places greater emphasis on activation directions that are more strongly expressed under the proxy distribution, thereby providing a stronger guidance signal for subspace decomposition. Accordingly, COMPASS adopts $\mathbf{C}$ as the proxy guidance matrix in the following decomposition.

\noindent \textbf{Covariance-Guided SVD.}
After selecting $\mathbf{C}$ as the proxy guidance matrix, COMPASS incorporates it directly into the weight decomposition. Specifically, we perform SVD on the covariance-guided weight matrix:
\begin{equation}
    \mathbf{W}\mathbf{C}
    =
    \mathbf{U}'\mathbf{S}'{\mathbf{V}'}^\top,
\end{equation}
where $\mathbf{U}'$ and $\mathbf{V}'$ contain the left and right singular vectors, respectively, and $\mathbf{S}'$ contains the singular values in descending order. The original weight matrix can be written as
\begin{equation}
    \mathbf{W}
    =
    \mathbf{U}'\mathbf{S}'{\mathbf{V}'}^\top
    \mathbf{C}^{-1}.
\end{equation}

Compared with standard SVD, whose component ordering is determined solely by the weight matrix, this covariance-guided decomposition also incorporates the activation statistics estimated from the proxy distribution. As a result, the resulting components reflect both the structure of the pre-trained weights and the target activation statistics estimated by the proxy set. These components are then divided into principal and residual parts for subsequent semantic anchoring and residual fine-tuning.

\subsection{Semantic Anchoring and Residual Forgery Learning}
\label{sec:compass_train}

Given the covariance-guided decomposition, we partition the singular components into principal and residual parts:
\begin{equation}
\mathbf{W}_{prin}
=
\mathbf{U}'_k \mathbf{S}'_k {\mathbf{V}'_k}^{\top} \mathbf{C}^{-1},
\qquad
\mathbf{W}_{res}
=
\mathbf{U}'_r \mathbf{S}'_r {\mathbf{V}'_r}^{\top} \mathbf{C}^{-1},
\end{equation}
where $k$ denotes the rank threshold. By construction,
$\mathbf{W}_{prin}+\mathbf{W}_{res}=\mathbf{W}$.

\noindent \textbf{Semantic Anchoring.}
We freeze $\mathbf{W}_{prin}$ throughout fine-tuning to preserve a stable representation of the dominant semantic content. The preserved principal component is not intended to serve as a direct cue for real/fake discrimination; instead, it provides a stable semantic reference within which forgery-related variations can be learned.

\noindent \textbf{Residual Forgery Learning.}
We introduce a trainable adapter $\Delta\mathbf{W}$ and initialize it with the residual component obtained from the covariance-guided decomposition:
\begin{equation}
\Delta\mathbf{W}^{(0)}
=
\mathbf{W}_{res}
=
\mathbf{U}'_r \mathbf{S}'_r {\mathbf{V}'_r}^{\top} \mathbf{C}^{-1}.
\end{equation}
Since $\mathbf{W}_{prin}+\Delta\mathbf{W}^{(0)}=\mathbf{W}$, the decomposed layer is exactly equivalent to the original pre-trained layer before optimization.

During fine-tuning, $\mathbf{W}_{prin}$ remains fixed and only the low-rank parameters of the residual adapter $\Delta\mathbf{W}$ are updated. The adapted layer is therefore
\begin{equation}
\mathbf{Y}
=
(\mathbf{W}_{prin}+\Delta\mathbf{W})\mathbf{X}.
\end{equation}
This residual adaptation maintains the principal weight component as a stable semantic reference while allowing the trainable component to capture forgery-related variations.

\noindent \textbf{Training Objective.}
Given the labeled source-domain training set $\mathcal{D}_{src}$, we optimize only $\Delta\mathbf{W}$ using the standard binary cross-entropy loss.

\section{Experiments}
\label{sec:experiments}

To empirically evaluate COMPASS, we conduct extensive experiments across diverse generalization settings. We first benchmark facial Deepfake detection against 16 methods under cross-dataset, cross-method, and joint cross-dataset-method protocols. We then extend the evaluation to open-world AIGI detection, comparing against 14 competitive detectors across 25 generative models. Finally, we perform a series of ablation and robustness studies to examine the calibration mechanism.

\subsection{Experimental Setups}
\noindent \textbf{Implementation Details.}
Following EFFORT~\citep{yan2025effort}, we use CLIP ViT-L/14~\citep{clip} as the VFM backbone. The unlabeled Semantic Proxy Set is sampled exclusively from the target-category training split and is used only to estimate activation statistics. No test samples are used for proxy-set, thereby avoiding test-set leakage despite the absence of label supervision. We use $N_p=80$ for Deepfake detection and $N_p=1500$ for AIGI detection unless otherwise specified.

During fine-tuning, only the residual adapter $\Delta\mathbf{W}$ is optimized. We use Adam with a learning rate of $2\times10^{-4}$ and batch sizes of 32 and 48 for Deepfake and AIGI detection, respectively. Following EFFORT, the residual rank is set to 1. All experiments are implemented in PyTorch and trained on a single NVIDIA RTX 4090 GPU.

\noindent \textbf{Evaluation Metrics.}
We report video-level AUC for Deepfake detection and AP for AIGI detection following AIGIBench~\citep{li2025artificial}.

\begin{table}[h]
\centering
\caption{Video-level AUC results of Cross-dataset and Cross-method evaluations. All detectors are trained on FF++ (c23) and evaluated on other fake data.}
\label{tab:deepfake_results}
\begin{adjustbox}{max width=\linewidth}
\begin{tabular}{l | ccccc | ccccccccc}
\toprule
\multirow{2}{*}{Methods} & \multicolumn{5}{c|}{Cross-dataset Evaluation} & \multicolumn{9}{c}{Cross-method Evaluation} \\
\cmidrule(lr){2-6} \cmidrule(lr){7-15}
& CDF-v1 & CDF-v2 & DFDC & DFDCP & Avg. & UniFace & BleFace & MobSwap & e4s & FaceDan & FSGAN & InSwap & SimSwap & Avg. \\
\midrule
F3Net~\citep{qian2020thinking}    & 0.777 & 0.789 & 0.718 & 0.749 & 0.758 & 0.809 & 0.808 & 0.867 & 0.494 & 0.717 & 0.845 & 0.757 & 0.674 & 0.746 \\
SPSL~\citep{liu2021spatial}       & 0.815 & 0.799 & 0.724 & 0.770 & 0.777 & 0.747 & 0.748 & 0.885 & 0.514 & 0.666 & 0.812 & 0.643 & 0.665 & 0.710 \\
SRM~\citep{liu2021spatial}        & 0.793 & 0.840 & 0.695 & 0.728 & 0.764 & 0.749 & 0.704 & 0.779 & 0.704 & 0.659 & 0.772 & 0.793 & 0.694 & 0.732 \\
CORE~\citep{ni2022core}           & 0.780 & 0.809 & 0.721 & 0.720 & 0.758 & 0.871 & 0.843 & 0.959 & 0.679 & 0.774 & 0.958 & 0.855 & 0.724 & 0.833 \\
RECCE~\citep{cao2022end}          & 0.768 & 0.823 & 0.696 & 0.734 & 0.755 & 0.898 & 0.832 & 0.925 & 0.683 & 0.848 & 0.949 & 0.848 & 0.768 & 0.844 \\
SLADD~\citep{chen2022ost}         & 0.802 & 0.837 & 0.772 & 0.756 & 0.792 & 0.878 & 0.882 & 0.954 & 0.765 & 0.825 & 0.943 & 0.879 & 0.794 & 0.865 \\
SBI~\citep{shiohara2022detecting} & 0.831 & 0.886 & 0.717 & 0.848 & 0.821 & 0.724 & 0.891 & 0.952 & 0.750 & 0.594 & 0.803 & 0.712 & 0.701 & 0.766 \\
UCF~\citep{yan2023ucf}            & 0.780 & 0.837 & 0.742 & 0.770 & 0.782 & 0.831 & 0.827 & 0.950 & 0.731 & 0.862 & 0.937 & 0.809 & 0.647 & 0.824 \\
IID~\citep{huang2023implicit}     & 0.758 & 0.838 & 0.700 & 0.689 & 0.746 & 0.839 & 0.789 & 0.888 & 0.766 & 0.844 & 0.927 & 0.789 & 0.644 & 0.811 \\
LSDA~\citep{yan2024transcending}  & 0.867 & 0.875 & 0.701 & 0.812 & 0.814 & 0.872 & 0.875 & 0.930 & 0.694 & 0.721 & 0.939 & 0.855 & 0.793 & 0.835 \\
ProDet~\citep{cheng2024can}       & 0.909 & 0.926 & 0.707 & 0.828 & 0.843 & 0.908 & 0.929 & 0.975 & 0.771 & 0.747 & 0.928 & 0.837 & 0.844 & 0.867 \\
CDFA~\citep{lin2024fake}          &   -   & 0.938 & 0.830 & 0.881 & 0.883 & 0.762 & 0.756 & 0.823 & 0.631 & 0.803 & 0.942 & 0.772 & 0.757 & 0.781 \\
Effort~\citep{yan2025effort}      & 0.970 & 0.956 & 0.843 & 0.909 & 0.920& 0.962 & 0.873 & 0.953 & 0.983 & 0.926 & 0.957 & 0.936 & 0.926 & 0.940 \\
\midrule
\rowcolor[HTML]{E8F0FE} 
Ours & 0.957 & 0.942 & 0.870 & 0.914 & 0.921 & 0.980 & 0.907 & 0.972 & 0.993 & 0.967 & 0.986 & 0.968 & 0.960 & 0.966 \\
\bottomrule
\end{tabular}
\end{adjustbox}
\end{table}

\subsection{Deepfake Detection}
Following the evaluation organization of EFFORT~\citep{yan2025effort}, we consider two established generalization settings: cross-dataset evaluation based on DeepfakeBench~\citep{yan2023deepfakebench} and cross-method evaluation based on DF40~\citep{yan2024df40}. We further introduce a joint cross-dataset-method setting that shifts both the data domain and manipulation method. We compare against 16 representative detectors; detailed baselines and datasets are summarized in \cref{tab:deepfake_results,tab:joint_results}.

\noindent \textbf{Evaluation Protocols.}
\begin{itemize}[leftmargin=*, labelindent=0pt, itemsep=1pt, topsep=2pt]
    \item \textbf{Protocol-1 (Cross-dataset):}
    Models trained on FF++ are evaluated on unseen facial datasets while keeping the forgery setting relatively consistent, measuring robustness to domain shift.

    \item \textbf{Protocol-2 (Cross-method):}
    Models are evaluated on unseen manipulation methods from DF40~\citep{yan2024df40} while keeping the underlying FF++ data domain fixed, isolating the effect of forgery-method shift. We report Face Swapping as the representative category in the main table.

    \item \textbf{Protocol-3 (Joint Cross-dataset-method):}
    Both the data domain and manipulation method are shifted simultaneously. Models trained on FF++ are evaluated on Face Reenactment forgeries generated from the CDF-v1 domain in DF40~\citep{yan2024df40}, providing a more challenging double-shift setting.
\end{itemize}

\noindent \textbf{Results and Analysis.}
As shown in \cref{tab:deepfake_results}, COMPASS achieves comparable or improved performance under both single-shift protocols. In cross-dataset evaluation, the gain over EFFORT is marginal (0.921 vs.\ 0.920 Avg.\ AUC). This setting changes the facial data domain while retaining relatively familiar manipulation characteristics, allowing both methods to continue exploiting forgery cues learned from the source distribution. In contrast, the advantage of COMPASS becomes more evident under cross-method evaluation, where the manipulation mechanism itself changes. Although the underlying FF++ domain remains familiar, preserving target-relevant pre-trained representations becomes more useful when previously learned forgery cues transfer less directly.

The advantage is largest under the joint cross-dataset-method setting in \cref{tab:joint_results}, where both the image domain and manipulation method are unseen. COMPASS improves the Avg.\ AUC from 0.710 for EFFORT to 0.829, an absolute gain of 11.9\%. These results suggest that the limitation of data-agnostic, magnitude-based preservation becomes increasingly consequential as fewer source-aligned cues remain available, explaining why the improvement is modest under cross-dataset evaluation but substantially larger under cross-method and joint shifts.

\begin{table}[h]
\centering
\caption{Video-level AUC results of Joint Cross-dataset-method evaluations. All detectors are trained on FF++ (c23) and evaluated on Face Reenactment forgeries generated from real CDF face data. All baselines are evaluated using the official checkpoints provided by their respective authors.}
\label{tab:joint_results}
\begin{adjustbox}{max width=\linewidth}
\begin{tabular}{l | ccccccccccccc}
\toprule
Methods & FOMM & FV2V & Wav2Lip & MRAA & OneShot & PiRender & TPSMM & LIA & DaGan & SadTalker & MCNet & HyperReenact & Avg. \\
\midrule
EFFNB4~\citep{tan2019efficientnet} & 0.799 & 0.795 & 0.683 & 0.880 & 0.833 & 0.759 & 0.620 & 0.772 & 0.602 & 0.555 & 0.683 & 0.637 & 0.718 \\
CNNAUG~\citep{wang2020cnn}         & 0.792 & 0.730 & 0.559 & 0.819 & 0.760 & 0.708 & 0.654 & 0.686 & 0.606 & 0.627 & 0.630 & 0.654 & 0.685 \\
F3Net~\citep{qian2020thinking}     & 0.689 & 0.741 & 0.726 & 0.927 & 0.660 & 0.648 & 0.552 & 0.836 & 0.575 & 0.452 & 0.663 & 0.608 & 0.673 \\
FFD~\citep{dang2020detection}      & 0.752 & 0.658 & 0.595 & 0.905 & 0.727 & 0.696 & 0.535 & 0.779 & 0.531 & 0.461 & 0.588 & 0.541 & 0.647 \\
SPSL~\citep{liu2021spatial}        & 0.769 & 0.773 & 0.697 & 0.851 & 0.742 & 0.742 & 0.633 & 0.681 & 0.583 & 0.539 & 0.635 & 0.800 & 0.704 \\
SRM~\citep{liu2021spatial}         & 0.800 & 0.810 & 0.750 & 0.937 & 0.804 & 0.729 & 0.665 & 0.856 & 0.652 & 0.559 & 0.735 & 0.651 & 0.746 \\
CORE~\citep{ni2022core}            & 0.724 & 0.678 & 0.680 & 0.930 & 0.714 & 0.659 & 0.601 & 0.853 & 0.517 & 0.493 & 0.620 & 0.684 & 0.679 \\
RECCE~\citep{cao2022end}           & 0.759 & 0.702 & 0.846 & 0.970 & 0.687 & 0.703 & 0.593 & 0.881 & 0.517 & 0.502 & 0.631 & 0.730 & 0.710 \\
UCF~\citep{yan2023ucf}             & 0.774 & 0.752 & 0.710 & 0.948 & 0.739 & 0.726 & 0.570 & 0.872 & 0.597 & 0.461 & 0.678 & 0.630 & 0.631 \\
Effort~\citep{yan2025effort}       & 0.798 & 0.709 & 0.636 & 0.841 & 0.755 & 0.729 & 0.619 & 0.797 & 0.611 & 0.540 & 0.638 & 0.843 & 0.710 \\
\midrule
\rowcolor[HTML]{E8F0FE} 
Ours                             & 0.919 & 0.853 & 0.715 & 0.970 & 0.906 & 0.873 & 0.724 & 0.925 & 0.771 & 0.563 & 0.813 & 0.919 & 0.829 \\
\bottomrule
\end{tabular}
\end{adjustbox}
\end{table}

\begin{table}[t]
\centering
\caption{Cross-generator evaluation on representative test sets spanning different generation families, together with the average performance over all 25 datasets included in AIGIBench~\citep{li2025artificial}. Full per-dataset results are provided in \cref{Full_Results_on_AIGIBench}.}
\label{tab:aigi_bench}
\begin{adjustbox}{max width=\linewidth}
\begin{tabular}{l | cccccccccccccc | >{\columncolor[HTML]{E8F0FE}}c}
\toprule
Generator & CNN-Spot & UnivFD & FreqNet & NPR & Ladeda & AIDE & C2P-CLIP & DFfreq & SAFE & Effort & NS-Net & FerretNet & LOTA & BSF & Ours \\
\midrule
R3GAN & 52.4 & 91.2 & 55.8 & 61.1 & 72.6 & 97.1 & 91.3 & 90.6 & 99.2 & 98.2 & 91.3 & 99.0 & 94.7 & 93.9 & 99.7 \\
StyleGAN-XL & 50.2 & 93.3 & 83.7 & 75.3 & 93.1 & 93.2 & 98.9 & 69.5 & 90.8 & 97.0 & 96.1 & 95.6 & 97.4 & 97.2 & 98.7 \\
FaceSwap & 58.3 & 52.3 & 42.9 & 43.6 & 40.9 & 56.3 & 70.4 & 64.9 & 69.6 & 85.8 & 62.5 & 71.6 & 73.4 & 66.4 & 88.6 \\
DALLE-3 & 68.4 & 76.4 & 60.2 & 70.0 & 59.8 & 63.1 & 56.3 & 48.5 & 49.5 & 60.6 & 42.2 & 55.4 & 44.3 & 43.0 & 92.3 \\
FLUX1-dev & 72.1 & 79.5 & 87.0 & 99.0 & 98.7 & 93.4 & 84.3 & 98.2 & 99.0 & 74.4 & 94.1 & 99.2 & 98.0 & 97.5 & 84.8 \\
SDXL & 63.8 & 88.0 & 95.0 & 94.4 & 98.5 & 95.7 & 97.1 & 99.1 & 99.4 & 98.0 & 98.2 & 99.9 & 99.4 & 98.0 & 94.4 \\
Infinite-ID & 49.1 & 89.6 & 73.6 & 80.4 & 76.9 & 94.7 & 84.2 & 97.4 & 98.5 & 85.7 & 91.8 & 99.7 & 98.1 & 97.0 & 97.2 \\
CommunityAI & 56.7 & 73.7 & 63.5 & 62.9 & 56.3 & 61.0 & 53.2 & 64.2 & 64.3 & 64.3 & 54.0 & 53.0 & 49.6 & 57.5 & 70.3 \\
\midrule
Average of all 25 & 67.3 & 75.7 & 69.9 & 73.9 & 79.3 & 82.7 & 82.2 & 82.2 & 82.9 & 87.2 & 83.0 & 85.8 & 82.1 & 81.1 & 90.9 \\
\bottomrule
\end{tabular}
\end{adjustbox}
\end{table}

\subsection{AI-Generated Image (AIGI) Detection}

For open-world AIGI detection, we follow Setting-II of AIGIBench~\citep{li2025artificial}. The setting uses 144K training images comprising real images and synthetic images from ProGAN~\citep{progan} and SDv1.4~\citep{stable-diffusion}, and evaluates generalization across 25 generators spanning diverse model families and generation paradigms.

We compare COMPASS with 14 representative AIGI detectors, covering CNN-, frequency-, VFM-based, and recent state-of-the-art methods. Detailed baseline information and references are provided in \cref{Full_Results_on_AIGIBench}.

As reported in \cref{tab:aigi_bench}, COMPASS achieves the best average AP of 90.9\% across 25 generative models. The improvement is particularly pronounced on several challenging commercial generators. For example, on DALLE-3~\citep{dalle3}, COMPASS reaches 92.3\% AP compared with 60.6\% for EFFORT, an absolute gain of 31.7\%. Consistent improvements are also observed on FLUX1-dev~\citep{flux2024} and CommunityAI. These results suggest that covariance-guided preservation remains effective under the substantially broader semantic and generative diversity of open-world AIGI detection, where magnitude-based decomposition may preserve less relevant pre-trained information.

\subsection{Ablation and Generalization Analysis}
\label{subsec:ablation}

\subsubsection{Robustness to Proxy Set Selection.}
\label{subsec:random_proxy}

We first examine whether COMPASS is sensitive to the datasets used to construct the Semantic Proxy Set. For each manipulation category, i.e., Face Swapping and Face Reenactment, we randomly select two datasets and sample a total of 80 unlabeled images from their training splits to form the proxy set. We repeat this procedure with different randomly selected dataset pairs and evaluate the resulting models across datasets within the same manipulation category.

As summarized in \cref{tab:proxy_robustness}(a), the performance remains stable across different proxy-source pairs. In FS, the average AUC ranges from 0.956 to 0.969 across four randomly selected pairs, compared with 0.939 for EFFORT. In FR, the corresponding range is 0.871--0.899 across six pairs, compared with 0.808 for EFFORT. These results indicate that COMPASS's proxy set does not depend strongly on particular datasets .




\begin{table}[t]
\centering
\caption{Robustness analysis of the Semantic Proxy Set.
(a) Sensitivity to randomly selected proxy-source datasets. Detailed results are provided in \cref{app:proxy_source}.
(b) Ablation on proxy composition, sample size, and semantic relevance.}
\label{tab:proxy_robustness}

\textbf{(a) Proxy-Source Selection}\par
\vspace{2pt}

\begin{adjustbox}{max width=0.70\linewidth}
\begin{tabular}{l c c c c}
\toprule
Category & EFFORT & COMPASS & Range & $\Delta$ \\
\midrule
Face Swapping
& 0.939
& \textbf{0.964 $\pm$ 0.005}
& 0.956--0.969
& +0.025 \\

Face Reenactment
& 0.808
& \textbf{0.888 $\pm$ 0.009}
& 0.871--0.899
& +0.080 \\
\bottomrule
\end{tabular}
\end{adjustbox}

\vspace{5pt}

\textbf{(b) Proxy-Set Ablation}\par
\vspace{2pt}

\begin{adjustbox}{max width=0.70\linewidth}
\begin{tabular}{l l c c c}
\toprule
Setting & Proxy Set & Ratio & $N_p$ & AUC (\%) \\
\midrule
Baseline
& EFFORT (No Proxy)
& -- & 0 & 94.0 \\

Default
& COMPASS (40 runs)
& Mixed & 80
& \textbf{95.8 $\pm$ 0.5} \\
\midrule

Composition
& All Real
& 100:0 & 80 & 95.76 \\

& All Fake
& 0:100 & 80 & 94.95 \\

Few-shot
& Two-shot
& 50:50 & 2 & 94.74 \\

& One-shot
& Random & 1 & 94.58 \\

Unrelated
& ImageNet/COCO
& -- & 80 & 93.62 \\
\bottomrule
\end{tabular}
\end{adjustbox}

\end{table}

\begin{figure}[h]
\centering

\pgfplotslegendfromname{sharedlegend}\par\medskip

\begin{minipage}[b]{0.47\linewidth}
\centering
\begin{tikzpicture}
\begin{axis}[
    width=\linewidth,
    height=4.3cm,
    xlabel={Number of Samples ($N_p$)},
    yticklabel style={
        /pgf/number format/fixed,
        /pgf/number format/fixed zerofill,
        /pgf/number format/precision=3
    },
    ylabel={Average AUC},
    xmin=0, xmax=320,
    ymin=0.935, ymax=0.970, 
    xtick={5, 40, 80, 160, 320},
    legend to name=sharedlegend, 
    legend columns=-1,           
    legend style={
        nodes={scale=0.85, transform shape}, 
        /tikz/every even column/.append style={column sep=0.5cm} 
    },
    ymajorgrids=true,
    grid style=dashed,
    thick
]

\addplot[color=red, dashed, mark=none, thick] 
coordinates {(0,0.940) (320,0.940)}; 
\addlegendentry{Effort}

\addplot[color=blue, mark=square*, thick] 
coordinates {
    (5,0.945) (10,0.951) (20,0.954) 
    (40,0.950) (80,0.966) (160,0.962) (320,0.963)
};
\addlegendentry{COMPASS}
\end{axis}
\end{tikzpicture}
\centerline{(a) Deepfake Detection}
\end{minipage}
\hfill 
\begin{minipage}[b]{0.47\linewidth}
\centering
\begin{tikzpicture}
\begin{axis}[
    width=\linewidth,
    height=4.3cm,
    xlabel={Number of Samples ($N_p$)},
    yticklabel style={
        /pgf/number format/fixed,
        /pgf/number format/fixed zerofill,
        /pgf/number format/precision=3
    },
    ylabel={Average AP}, 
    xmin=0, xmax=80,
    ymin=0.870, ymax=0.890, 
    xtick={5, 20, 40, 60, 80},                           
    xticklabels={125, 500, 1000, 1500, 2000},            
    ymajorgrids=true,
    grid style=dashed,
    thick
]

\addplot[color=red, dashed, mark=none, thick] 
coordinates {(0,0.872) (80,0.872)}; 

\addplot[color=blue, mark=square*, thick] 
coordinates {
    (5,0.872) (10,0.872) (15,0.873) 
    (20,0.874) (30,0.884) (40,0.883) (60,0.890) (80, 0.886)
};
\end{axis}
\end{tikzpicture}
\centerline{(b) AIGI Detection}
\end{minipage}

\caption{Sensitivity to Proxy Set Size ($N_p$). COMPASS improves with additional proxy set samples and gradually stabilizes on both (a) Deepfake Detection and (b) AIGI Detection. The facial domain requires substantially fewer samples than the more semantically diverse open-world AIGI setting.}
\label{fig:ablation_nc}
\end{figure}
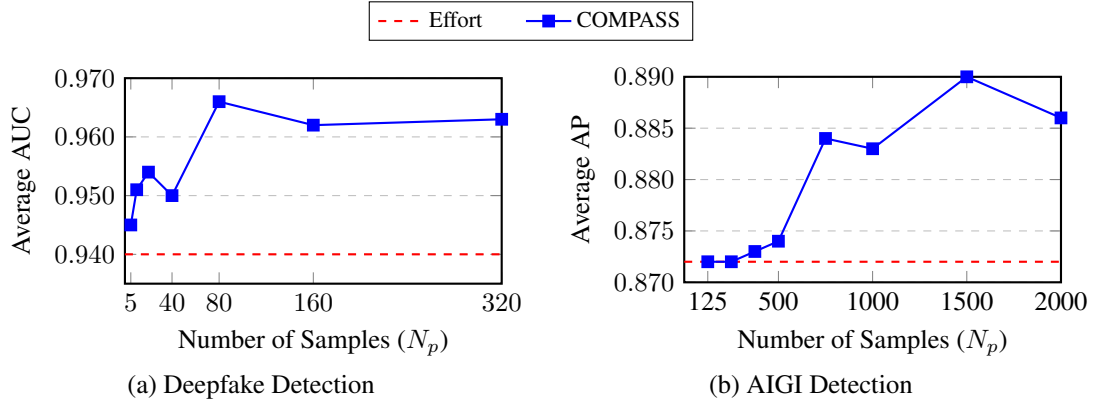

We further vary the composition and size of the proxy set in \cref{tab:proxy_robustness}(b). Across 40 random proxy sets, COMPASS achieves an average AUC of 95.8\% with a standard deviation of 0.5\%. The performance remains above the EFFORT baseline when using only real or only fake samples, and even one- or two-sample proxy sets still provide a modest improvement. In contrast, using unrelated ImageNet/COCO images reduces performance to 93.62\%, close to the EFFORT baseline. These results indicate that COMPASS benefits from a small amount of target-relevant semantic context, while being relatively insensitive to the precise real/fake composition of the proxy set.

\subsubsection{Effect of Proxy Set Size.}
\label{subsec:ablation_nc}

As shown in \cref{fig:ablation_nc}, COMPASS is effective with a small proxy set and stabilizes as $N_p$ increases. For Deepfake detection, performance peaks at $N_p=80$ (0.966 AUC), while the more diverse AIGI setting requires more samples, reaching 0.890 AP at $N_p=1500$. This difference likely reflects the broader semantic diversity of open-world AIGI detection compared with the relatively constrained facial domain. Even so, 1,500 unlabeled samples represent only a small fraction of the 100K+ training corpus. These results suggest that COMPASS can estimate useful target activation statistics from comparatively limited unlabeled data, although the required proxy-set size increases with the diversity of the target domain.

\subsubsection{Guidance Matrix Design.}
\label{subsec:StatisticSelection}

We compare different statistics for constructing the proxy guidance matrix in \cref{tab:ablation_generalization}(a). Overall, the full covariance $\mathbf{C}$ achieves the best performance among the evaluated variants, outperforming mean-, correlation-, and diagonal-variance-based guidance on both Deepfake and AIGI detection.

Notably, although $\mathbf{C}^{1/2}$ follows naturally from the response-preservation objective in \cref{sec:compass_formulation}, it yields 0.955 AUC and 0.887 AP, compared with 0.966 and 0.890 using $\mathbf{C}$. We therefore adopt the full covariance as an empirical design choice, suggesting that the stronger covariance weighting provides more effective guidance for data-aware decomposition.

\subsubsection{Generalization across VFM Backbones.}
We evaluate COMPASS with BEiT-v2, SigLIP, DINOv3 and CLIP under the cross-method setting. As shown in \cref{tab:ablation_generalization}(b), COMPASS improves all three backbones by 0.125--0.190 AUC, suggesting that proxy-guided subspace calibration is not specific to CLIP or a single pre-training paradigm.

\begin{table}[t]
\centering
\caption{
(a) Comparison of different proxy guidance matrices on Deepfake and AIGI detection.
(b) Cross-method evaluation across different VFM backbones.}
\label{tab:ablation_generalization}

\textbf{(a) Proxy Guidance Matrix}\par
\vspace{2pt}

\begin{adjustbox}{max width=0.60\linewidth}
\begin{tabular}{l l c c}
\toprule
Method & Guidance Matrix $\mathbf{G}$ 
& Deepfake AUC $\uparrow$ 
& AIGI AP $\uparrow$ \\
\midrule
EFFORT
& Identity ($\mathbf{I}$)
& 0.940 & 0.872 \\
\midrule

Mean
& Rank-1 Mean ($\bar{\mathbf{X}}_p\bar{\mathbf{X}}_p^\top$)
& 0.943 & 0.873 \\

Raw
& Raw Correlation ($\frac{1}{N_p}\mathbf{X}_p\mathbf{X}_p^\top$)
& 0.956 & 0.879 \\

Diag
& Diagonal Variance ($\operatorname{diag}(\mathbf{C})$)
& 0.938 & 0.874 \\

Sqrt
& Covariance Root ($\mathbf{C}^{1/2}$)
& 0.955 & 0.879 \\
\midrule

\rowcolor[HTML]{E8F0FE}
COMPASS
& Covariance ($\mathbf{C}$)
& \textbf{0.966} & \textbf{0.890} \\
\bottomrule
\end{tabular}
\end{adjustbox}

\vspace{5pt}

\textbf{(b) VFM Backbone Generalization}\par
\vspace{2pt}

\begin{adjustbox}{max width=0.62\linewidth}
\begin{tabular}{l l c c c}
\toprule
Backbone & Pre-training & Base AUC & +COMPASS & $\Delta$ \\
\midrule
BEiT-v2~\citep{peng2022beit}
& MIM
& 0.659
& \textbf{0.824}
& \textbf{+0.165} \\

SigLIP~\citep{zhai2023sigmoid}
& Contrastive
& 0.714
& \textbf{0.839}
& \textbf{+0.125} \\

DINOv3~\citep{simeoni2025dinov3}
& Self-supervised
& 0.772
& \textbf{0.962}
& \textbf{+0.190} \\

CLIP~\citep{clip}
& Self-supervised
& 0.851
& \textbf{0.966}
& \textbf{+0.115} \\
\bottomrule
\end{tabular}
\end{adjustbox}

\end{table}

\noindent \textbf{Robustness to Input Perturbations.}
We additionally evaluate JPEG compression, Gaussian blur, Gaussian noise, and resizing. COMPASS maintains higher AUC than EFFORT under all tested perturbations, with an average AUC of 0.948 versus 0.927. Detailed results are provided in \cref{real_world_perturbations}.

\section{Conclusion}
We present COMPASS, a data-aware subspace adaptation framework for AI-generated image and deepfake detection. Using a small unlabeled Semantic Proxy Set, COMPASS estimates target activation statistics to guide pre-trained weight decomposition, preserving target-relevant representations as a stable semantic reference while the residual adapter learns forgery-related variations. Experiments across cross-dataset, cross-method, joint cross-dataset-method, and open-world AIGI settings demonstrate strong generalization, with particularly large gains under simultaneous domain and manipulation shifts. Ablations further show robustness to proxy selection, proxy set size, and different VFM backbones. Overall, our results support covariance-guided subspace preservation as an effective strategy for generalizable forgery detection.


\bibliography{iclr2027_conference}
\bibliographystyle{iclr2027_conference}

\clearpage
\appendix
\section{Appendix}

\subsection{Mathematical Detail: Systematic Scaling Offset in Feature Similarity}
\label{sec:math_offset}

As observed in the pilot study (\cref{fig:pilot}(b) of the main paper), the cosine similarities in the compressed subspace systematically shift above the $y=x$ diagonal. We provide a geometric explanation for this similarity shift under principal-subspace projection. The analysis proceeds in three stages.

\subsubsection{Concentration of Measure in High-Dimensional Space}
We first establish the baseline behavior of feature vectors in a high-dimensional space $\mathbb{R}^d$. Let $\mathbf{u}, \mathbf{v} \in \mathbb{R}^d$ be two randomly distributed feature vectors. For simplicity, assume their elements are independent and identically distributed (i.i.d.) random variables with zero mean and variance $\sigma^2$. 

The expected inner product of these vectors is:
\begin{equation}
\mathbb{E}[\langle \mathbf{u}, \mathbf{v} \rangle] = \mathbb{E}\left[\sum_{i=1}^d u_i v_i\right] = \sum_{i=1}^d \mathbb{E}[u_i]\mathbb{E}[v_i] = 0
\end{equation}
The variance of the inner product is $d\sigma^4$, implying its magnitude is approximately $\mathcal{O}(\sqrt{d}\sigma^2)$.

According to the Law of Large Numbers, the squared $\ell_2$-norm of $\mathbf{u}$ is:
\begin{equation}
\|\mathbf{u}\|^2 = \sum_{i=1}^d u_i^2 \approx d\sigma^2 \implies \|\mathbf{u}\| \approx \sqrt{d}\sigma
\end{equation}

Consequently, the magnitude of the original cosine similarity $S_{orig}$ is:
\begin{equation}
S_{orig} = \cos(\theta) = \frac{\langle \mathbf{u}, \mathbf{v} \rangle}{\|\mathbf{u}\| \|\mathbf{v}\|} \sim \mathcal{O}\left(\frac{\sqrt{d}\sigma^2}{\sqrt{d}\sigma \cdot \sqrt{d}\sigma}\right) = \mathcal{O}\left(\frac{1}{\sqrt{d}}\right)
\end{equation}
When $d$ is exceptionally large (e.g., $d=1024$), $1/\sqrt{d} \approx 0.031$. This illustrates the \textit{concentration of measure} phenomenon: in high-dimensional spaces, random vectors are highly likely to be nearly orthogonal, clustering their cosine similarities near zero.

\subsubsection{Orthogonal Decomposition via SVD Truncation}
When we apply SVD truncation to retain the top $k$ principal components, we project the original vectors into a lower-dimensional principal subspace. This acts as an orthogonal decomposition:
\begin{equation}
\mathbf{u} = \mathbf{u}_k + \mathbf{u}_{\perp}, \quad \mathbf{v} = \mathbf{v}_k + \mathbf{v}_{\perp}
\end{equation}
where $\mathbf{u}_k$ represents the retained principal semantics and $\mathbf{u}_{\perp}$ represents the discarded residual components (predominantly high-frequency noise). 

By the Pythagorean theorem, the squared norm satisfies:
\begin{equation}
\|\mathbf{u}\|^2 = \|\mathbf{u}_k\|^2 + \|\mathbf{u}_{\perp}\|^2
\end{equation}
This strictly dictates that the norm decreases after projection: $\|\mathbf{u}_k\| < \|\mathbf{u}\|$.

\subsubsection{Mechanism of Similarity Inflation}
We can now compare the cosine similarity before and after compression. The original similarity can be expanded as:
\begin{equation}
S_{orig} = \frac{\langle \mathbf{u}_k, \mathbf{v}_k \rangle + \langle \mathbf{u}_{\perp}, \mathbf{v}_{\perp} \rangle}{\|\mathbf{u}\| \|\mathbf{v}\|}
\end{equation}
The similarity in the compressed subspace is:
\begin{equation}
S_{comp} = \frac{\langle \mathbf{u}_k, \mathbf{v}_k \rangle}{\|\mathbf{u}_k\| \|\mathbf{v}_k\|}
\end{equation}

The core rationale of SVD is to capture directions of maximum variance. Therefore, the shared semantic structures between similar images are predominantly preserved within $\mathbf{u}_k$ and $\mathbf{v}_k$. Conversely, the discarded residual dimensions $\mathbf{u}_{\perp}$ and $\mathbf{v}_{\perp}$ act essentially as uncorrelated high-dimensional noise. 

Referring back to our first stage, the inner product of such uncorrelated high-dimensional noise vectors approaches zero: $\langle \mathbf{u}_{\perp}, \mathbf{v}_{\perp} \rangle \approx 0$. Consequently, the numerator (inner product) remains largely unchanged after truncation:
\begin{equation}
\langle \mathbf{u}, \mathbf{v} \rangle \approx \langle \mathbf{u}_k, \mathbf{v}_k \rangle
\end{equation}

However, the denominator (product of norms) strictly and significantly decreases due to dimensionality reduction ($\|\mathbf{u}_k\| \|\mathbf{v}_k\| < \|\mathbf{u}\| \|\mathbf{v}\|$). In a fraction, preserving the numerator while decreasing the denominator can lead to an increase in the overall value:
\begin{equation}
S_{comp} = \frac{\langle \mathbf{u}_k, \mathbf{v}_k \rangle}{\|\mathbf{u}_k\| \|\mathbf{v}_k\|} \approx \frac{\langle \mathbf{u}, \mathbf{v} \rangle}{\|\mathbf{u}_k\| \|\mathbf{v}_k\|} > \frac{\langle \mathbf{u}, \mathbf{v} \rangle}{\|\mathbf{u}\| \|\mathbf{v}\|} = S_{orig}
\end{equation}

\textbf{Conclusion \& Empirical Alignment:} The systematic scaling and distribution of cosine similarities in \cref{fig:pilot}(b). perfectly align with our mathematical derivation regarding subspace truncation. 

\textbf{First}, the similarity points in COMPASS are more concentrated (Spearman rank correlation $\rho = 0.5533$) compared to the broader scatter of EFFORT~\citep{yan2025effort} ($\rho = 0.1345$). The higher rank correlation suggests that COMPASS better preserves the relative similarity structure of the original space. By utilizing covariance-guided calibration, COMPASS better preserves the pre-trained representation, perfectly preserving the meaningful numerator (shared semantics $\langle \mathbf{u}_k, \mathbf{v}_k \rangle$) without disrupting structural relationships.

\textbf{Second}, the upward drift of similarities in COMPASS is more pronounced. Magnitude-based SVD prioritizes components according to their singular values. Its reduction of the vector magnitude (the denominator) is inherently limited, leading to a less-shifted but collapsed representation. In contrast, COMPASS discards components based on semantic irrelevance. It intentionally strips away ``magnitude-heavy but semantically useless'' domain biases. By discarding a much larger proportion of absolute magnitude (reducing the feature norms more strongly), COMPASS can therefore produce a larger upward shift in cosine similarity after normalization.

\subsection{Qualitative Analysis via Grad-CAM}
\label{sec:gradcam_analysis}

\begin{figure}[htbp]
  \centering
  \includegraphics[width=0.5\textwidth]{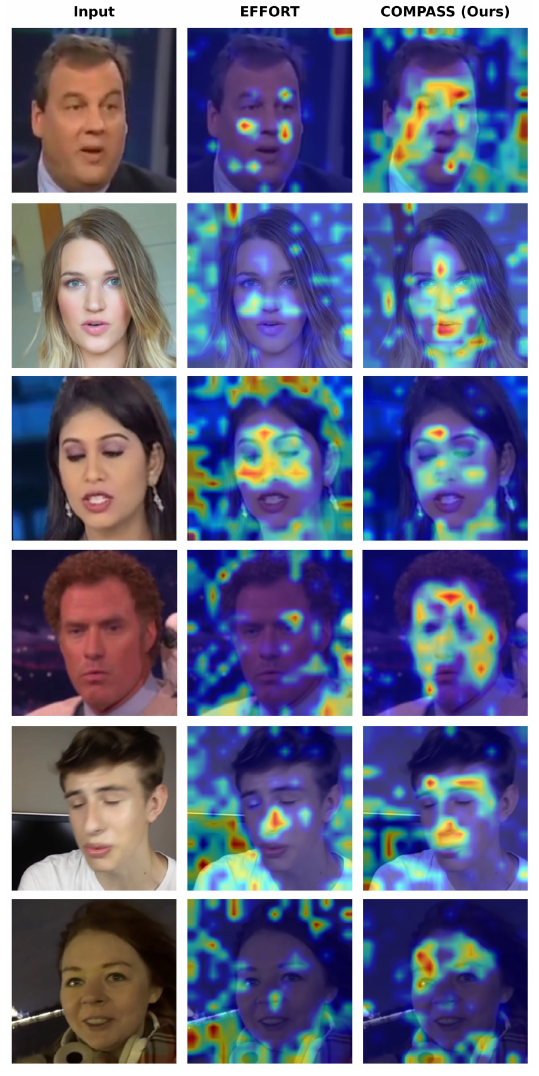}
  \caption{\textbf{Grad-CAM Visualization on Deepfake Detection.} Left: Input fake image. Middle: EFFORT exhibits relatively dispersed attention across facial and contextual regions. Right: COMPASS places more attention on the central facial region.}
  \label{fig:gradcam}
  \vspace{-6mm}
\end{figure}

To qualitatively examine the spatial attention patterns produced by EFFORT and COMPASS, we visualize Grad-CAM responses on a representative fake sample. As shown in \cref{fig:gradcam}, EFFORT exhibits a relatively dispersed attention pattern that extends to background and clothing regions in addition to the face. In contrast, COMPASS places more attention on the central facial region, including the eyes, nose, and mouth boundaries.

This observation is consistent with our motivation that better preservation of the pre-trained semantic reference can reduce reliance on source-specific contextual cues during residual adaptation. We emphasize that Grad-CAM provides only qualitative evidence; the generalization benefits of COMPASS are established by the quantitative cross-dataset and cross-method experiments in the main paper.

\subsection{Algorithm Overview}
\label{sec:algorithm_pseudocode}

To provide a clearer step-by-step implementation perspective, we summarize the entire workflow of the proposed COMPASS framework in Algorithm \ref{alg:compass}. Unlike static orthogonal methods that rely on heavy constraint losses during optimization, COMPASS performs proxy-guided subspace calibration prior to fine-tuning. Furthermore, by optimizing only the decomposed low-rank factors of the residual space rather than the full parameter matrix, it inherently maintains a structural bottleneck, preventing overfitting to domain biases.

\begin{algorithm}[t]
\caption{\textit{COMPASS} Approach Algorithm}
\label{alg:compass}
\textbf{Input:} Pre-trained weight matrix $W \in \mathbb{R}^{d_{out} \times d_{in}}$; Rank $k$; Training data $\mathcal{D}_{src} = \{(x_i, y_i)\}_{i=1}^M$; Unlabeled calibration set $\mathcal{D}_{p}$ of size $N_p$\\
\textbf{Output:} Updated weight matrix $W$
\begin{algorithmic}[1]
\STATE $\triangleright$ \textbf{Step 1: Proxy-Guided Subspace Calibration (Zero-Backpropagation)}
\STATE Pass $\mathcal{D}_{cal}$ through the frozen VFM to extract activations $X_p \in \mathbb{R}^{d_{in} \times N_p}$
\STATE Mean-center the activations: $\hat{X}_p = X_p - \bar{X}_p$
\STATE Compute the activation covariance: $C = \frac{1}{N_p}\hat{X}_p\hat{X}_p^\top$
\STATE $\triangleright$ \textbf{Step 2: Covariance-Guided SVD}
\STATE Decompose proxy-scaled weights via SVD: $WC = U'S'{V'}^\top$
\STATE Retain the top $k$ singular values and vectors: $U'_k \in \mathbb{R}^{d_{out} \times k}, S'_k \in \mathbb{R}^{k \times k}, V'_k \in \mathbb{R}^{d_{in} \times k}$
\STATE Compute the principal weights (Semantic Compass): $W_{prin} = U'_k S'_k {V'_k}^\top C^{-1}$
\STATE Keep $W_{prin}$ strictly fixed during training
\STATE Obtain the residual components for the adapter: $U'_r, S'_r, V'_r$
\STATE Initialize trainable low-rank factors: 
\STATE \quad $U_{train} \leftarrow U'_r$
\STATE \quad $S_{train} \leftarrow S'_r$
\STATE \quad $V_{train} \leftarrow {V'_r}^\top C^{-1}$
\STATE $\triangleright$ \textbf{Step 3: Training Loop}
\FOR{each epoch}
    \FOR{each batch in $\mathcal{D}_{src}$}
        \STATE $\triangleright$ Forward Pass
        \STATE Compute layer output using $Y = (W_{prin} + U_{train} S_{train} V_{train})X_{src}$
        \STATE Compute the binary cross-entropy loss $\mathcal{L}_{BCE}$
        \STATE $\triangleright$ Backward Pass and Optimization
        \STATE Update \textbf{only} $U_{train}, S_{train},$ and $V_{train}$ using gradient descent to minimize $\mathcal{L}_{CE}$
    \ENDFOR
\ENDFOR
\STATE \textbf{return} Updated weight matrix $W \leftarrow W_{prin} + U_{train} S_{train} V_{train}$
\end{algorithmic}
\end{algorithm}

\subsection{Detailed Implementation and Reproducibility}
\label{sec:implementation}

To ensure full reproducibility, we provide comprehensive implementation details of the proposed COMPASS framework across two diverse domains.

\textbf{Architecture \& Hardware:} Consistent with the baseline EFFORT, we adopt the pre-trained CLIP ViT-L/14~\citep{clip} as our Vision Foundation Model (VFM) backbone. To accommodate the different scales of the training sets, the deepfake detection models are trained on a single NVIDIA RTX 4090 GPU (24GB VRAM), while the large-scale AIGI detection models are trained on a single NVIDIA RTX A6000 GPU (48GB VRAM).

\textbf{Optimization \& Hyperparameters:} During the decoupled forgery feature learning phase, the prior-aligned principal weights $W_{prin}$ remain strictly frozen to preserve the principal weight component as a stable semantic reference. We exclusively optimize the low-rank residual adapter. To maintain parameter efficiency and prevent overfitting to domain-specific biases, the residual adapter rank is set to 1. The model is optimized using the Adam optimizer. For the deepfake video detection training phase, the batch size is set to 32 with a learning rate of $2 \times 10^{-4}$. For the large-scale open-world AIGI detection task, the batch size is set to 48, with the learning rate adjusted to $1 \times 10^{-4}$. 

\textbf{Calibration Phase:} The Semantic Proxy Set is constructed using randomly sampled, unlabeled images from the target-category training distribution. The covariance-guided Covariance-Guided SVD is performed entirely offline prior to the training phase, requiring zero backpropagation and introducing no gradient optimization overhead.

\subsection{Computational Cost and Complexity Analysis}
\label{sec:complexity}

A primary advantage of COMPASS is its exceptional parameter efficiency and negligible computational overhead compared to full fine-tuning paradigms.

\textbf{Parameter Efficiency:} By restricting the trainable parameters exclusively to the low-rank residual adapter, COMPASS drastically reduces the optimization burden. Based on our model structure, the trainable parameters account for only \textbf{0.38\%} of the full ViT-L/14 backbone (specifically, approximately 1.18M trainable parameters out of a total 304M parameters).

\textbf{Training Overhead \& Memory Footprint:} The proxy-guided subspace calibration step is a highly lightweight, one-time offline process. Extracting features for the minimal proxy set and computing the covariance-guided SVD takes only a few seconds on a single GPU. During the fine-tuning phase, thanks to the low-rank structural bottleneck, COMPASS operates with remarkable resource efficiency:
\begin{itemize}
    \item \textbf{Deepfake Detection:} Operating on the RTX 4090 with a batch size of 32, the peak GPU memory utilization is tightly constrained to approximately \textbf{15 GB}. One full epoch is processed rapidly in roughly \textbf{16 minutes}.
    \item \textbf{AIGI Detection:} Handling the significantly larger AIGIBench on the RTX A6000 with a batch size of 48, the peak GPU memory footprint is maintained at around \textbf{22 GB}. Despite the massive scale of the data, one full epoch requires only about \textbf{87 minutes}.
\end{itemize}

\textbf{Inference Complexity (FLOPs):} Since the trainable adapter $\Delta W$ and the frozen principal weights $W_{prin}$ are strictly parallel linear structures, they can be seamlessly re-parameterized into a single weight matrix $W_{deploy} = W_{prin} + \Delta W$ during the deployment phase. Consequently, COMPASS introduces absolutely \textbf{zero} additional inference latency or FLOPs overhead compared to the base zero-shot CLIP model, making it exceptionally well-suited for real-time, real-world generative artifact detection scenarios.

\subsection{Full Results on AIGIBench}
\label{Full_Results_on_AIGIBench}
We conduct experiments on the comprehensive AIGIBench~\citep{li2025artificial}. We adopt the standard Setting-II Protocol~\citep{li2025artificial} of AIGIBench, which includes 144k images generated by ProGAN~\citep{progan} and SDv1.4~\citep{stable-diffusion}, along with real images, to ensure a rigorous assessment. This evaluation encompasses 25 diverse generative models, ranging from GANs (e.g., ProGAN~\citep{progan}, StyleGAN3~\citep{stylegan3}) to the latest advanced Diffusion Models (e.g., Midjourney-V6~\citep{midjourney}, DALLE-3~\citep{dalle3}, SD3~\citep{esser2024scaling}, SDXL~\citep{podell2023sdxl}).

We compare our approach against 14 competitive detectors. This includes widely recognized methods such as CNN-Spot~\citep{wang2020cnn}, UnivFD~\citep{ojha2023towards}, FreqNet~\citep{tan2024frequency}, NPR~\citep{tan2024rethinking}, Ladeda~\citep{cavia2024real}, AIDE~\citep{yan2024sanity}, C2P-CLIP~\citep{tan2025c2p}, DFfreq~\citep{yan2026dual}, SAFE~\citep{li2025improving}, EFFORT~\citep{yan2025effort}; and the latest SOTA detectors including NS-Net~\citep{yan2025ns}, FerretNet~\citep{liang2025ferretnet}, LOTA~\citep{wang2025lota}, and BSF~\citep{zhou2025beyond}.

\cref{tab:aigi_full_results} reports the complete per-generator AP results for COMPASS and 14 representative AIGI detectors. 
While the main paper presents a subset of representative generators for compactness, the full results show that COMPASS maintains strong overall generalization across the diverse generator set.

\begin{table}[t]
\centering
\caption{AP Results on AIGIBench. All detectors are trained on the Setting-II of AIGIBench.}
\label{tab:aigi_full_results}
\resizebox{\textwidth}{!}{
\begin{tabular}{l | cccccccccccccc | >{\columncolor[HTML]{E8F0FE}}c}
\toprule
Generator & CNN-Spot & UnivFD & FreqNet & NPR & Ladeda & AIDE & C2P-CLIP & DFfreq & SAFE & Effort & NS-Net & FerretNet & LOTA & BSF & Ours \\
\midrule
ProGAN & 99.9 & 99.9 & 100.0 & 100.0 & 100.0 & 99.6 & 100.0 & 100.0 & 100.0 & 100.0 & 100.0 & 100.0 & 99.9 & 100.0 & 100.0 \\
R3GAN & 52.4 & 91.2 & 55.8 & 61.1 & 72.6 & 97.1 & 91.3 & 90.6 & 99.2 & 98.2 & 91.3 & 99.0 & 94.7 & 93.9 & 99.7 \\
StyleGAN3 & 85.2 & 84.5 & 92.4 & 91.7 & 96.9 & 91.4 & 99.5 & 98.1 & 94.7 & 98.9 & 99.4 & 92.2 & 96.5 & 95.8 & 99.5 \\
StyleGAN-XL & 50.2 & 93.3 & 83.7 & 75.3 & 93.1 & 93.2 & 98.9 & 69.5 & 90.8 & 97.0 & 96.1 & 95.6 & 97.4 & 97.2 & 98.7 \\
StyleSwim & 76.2 & 95.2 & 91.6 & 94.9 & 98.5 & 89.3 & 99.6 & 99.5 & 96.7 & 99.1 & 97.8 & 99.8 & 96.4 & 98.6 & 99.9 \\
WFIR & 65.5 & 83.0 & 57.0 & 65.5 & 86.9 & 90.8 & 99.6 & 89.0 & 60.7 & 100.0 & 99.4 & 98.1 & 78.8 & 51.5 & 98.6 \\
BlendFace & 73.2 & 35.2 & 34.0 & 34.7 & 42.1 & 54.2 & 48.9 & 42.3 & 44.1 & 66.1 & 54.4 & 51.7 & 45.9 & 47.9 & 68.9 \\
E4S & 68.7 & 57.0 & 34.5 & 34.4 & 49.3 & 44.3 & 48.3 & 41.0 & 45.6 & 76.9 & 49.4 & 42.6 & 44.1 & 38.4 & 79.1 \\
FaceSwap & 58.3 & 52.3 & 42.9 & 43.6 & 40.9 & 56.3 & 70.4 & 64.9 & 69.6 & 85.8 & 62.5 & 71.6 & 73.4 & 66.4 & 88.6 \\
InSwap & 77.5 & 40.1 & 41.8 & 40.7 & 47.4 & 54.6 & 65.6 & 57.3 & 62.3 & 83.1 & 67.5 & 59.6 & 52.4 & 49.5 & 83.2 \\
SimSwap & 69.7 & 40.3 & 41.7 & 42.7 & 42.3 & 62.7 & 68.5 & 57.7 & 59.9 & 89.7 & 69.5 & 68.7 & 55.7 & 62.9 & 88.6 \\
DALLE-3 & 68.4 & 76.4 & 60.2 & 70.0 & 59.8 & 63.1 & 56.3 & 48.5 & 49.5 & 60.6 & 42.2 & 55.4 & 44.3 & 43.0 & 92.3 \\
FLUX1-dev & 72.1 & 79.5 & 87.0 & 99.0 & 98.7 & 93.4 & 84.3 & 98.2 & 99.0 & 74.4 & 94.1 & 99.2 & 98.0 & 97.5 & 84.8 \\
Midjourney-V6 & 59.8 & 61.5 & 55.9 & 76.9 & 86.9 & 83.0 & 84.6 & 94.2 & 94.5 & 90.0 & 90.0 & 96.5 & 91.0 & 86.2 & 91.4 \\
GLIDE & 59.7 & 80.2 & 76.5 & 94.3 & 95.0 & 97.7 & 94.9 & 97.7 & 96.1 & 96.5 & 96.1 & 99.7 & 93.9 & 98.4 & 97.2 \\
Imagen3 & 57.0 & 79.3 & 80.2 & 94.4 & 97.2 & 95.2 & 93.9 & 84.7 & 97.0 & 97.8 & 95.9 & 97.3 & 88.5 & 94.0 & 96.9 \\
SD3 & 72.7 & 87.2 & 81.9 & 97.2 & 98.7 & 98.3 & 97.1 & 97.0 & 99.3 & 97.9 & 98.9 & 98.8 & 95.1 & 96.4 & 90.6 \\
SDXL & 63.8 & 88.0 & 95.0 & 94.4 & 98.5 & 95.7 & 97.1 & 99.1 & 99.4 & 98.0 & 98.2 & 99.9 & 99.4 & 98.0 & 94.4 \\
BLIP & 92.8 & 95.8 & 99.8 & 100.0 & 99.9 & 99.5 & 100.0 & 100.0 & 100.0 & 100.0 & 100.0 & 100.0 & 99.9 & 100.0 & 99.9 \\
Infinite-ID & 49.1 & 89.6 & 73.6 & 80.4 & 76.9 & 94.7 & 84.2 & 97.4 & 98.5 & 85.7 & 91.8 & 99.7 & 98.1 & 97.0 & 97.2 \\
PhotoMaker & 57.8 & 72.2 & 74.2 & 53.4 & 90.7 & 95.6 & 69.7 & 99.0 & 96.9 & 69.2 & 67.8 & 99.1 & 97.3 & 95.5 & 96.7 \\
Instant-ID & 79.9 & 93.5 & 86.0 & 79.2 & 90.4 & 96.3 & 84.9 & 99.9 & 94.8 & 87.1 & 87.7 & 100.0 & 98.5 & 99.0 & 98.8 \\
IP-Adapter & 65.5 & 87.3 & 79.4 & 91.7 & 94.3 & 95.4 & 95.4 & 98.8 & 94.8 & 98.1 & 96.2 & 99.9 & 97.6 & 98.7 & 91.3 \\
SocialRF & 50.6 & 55.2 & 58.1 & 68.4 & 68.3 & 65.0 & 67.7 & 66.9 & 63.5 & 65.2 & 74.6 & 68.0 & 65.0 & 64.0 & 65.2 \\
CommunityAI & 56.7 & 73.7 & 63.5 & 62.9 & 56.3 & 61.0 & 53.2 & 64.2 & 64.3 & 64.3 & 54.0 & 53.0 & 49.6 & 57.5 & 70.3 \\
\midrule
Average & 67.3 & 75.7 & 69.9 & 73.9 & 79.3 & 82.7 & 82.2 & 82.2 & 82.9 & 87.2 & 83.0 & 85.8 & 82.1 & 81.1 & 90.9 \\
\bottomrule
\end{tabular}
}
\end{table}

\subsection{Detailed Results on Proxy-Source Selection}
\label{app:proxy_source}

For completeness, we report the full per-dataset results corresponding to the proxy-source robustness analysis in \cref{tab:proxy_robustness}(a). For each manipulation category, we randomly select two datasets and sample a total of 80 \textbf{unlabeled} images from their training splits to construct the Semantic Proxy Set. The proxy images are used only to estimate activation statistics, and no test images are involved in calibration. We repeat this procedure with different dataset pairs and evaluate the resulting models on all datasets within the same manipulation category.

As shown in \cref{tab:random_proxy_results_full}, COMPASS remains consistently effective across different proxy-source pairs. For Face Swapping, the average AUC ranges from 0.956 to 0.969 across four proxy pairs, compared with 0.939 for EFFORT. For Face Reenactment, the corresponding range is 0.871--0.899 across six proxy pairs, compared with 0.808 for EFFORT. The detailed results support the conclusion in the main paper that COMPASS is relatively insensitive to the particular datasets used to construct a target-relevant proxy set.

\begin{table}[t]
\centering
\caption{
Detailed robustness analysis of proxy-source selection. For each Face Swapping (FS) or Face Reenactment (FR) setting, two datasets are selected as proxy sources and a total of 80 \textbf{unlabeled} images are sampled from their training splits. We report per-dataset AUC for different proxy-source pairs together with representative baselines.
}
\label{tab:random_proxy_results_full}

{(a) Face Swapping (FS) Category}\par

\vspace{0.05in}
\begin{adjustbox}{max width=\linewidth}
\begin{tabular}{l | c c c | c c c c}
\toprule
\multirow{2}{*}{Testing Dataset} & \multicolumn{3}{c|}{Baseline} & \multicolumn{4}{c}{Ours (Calibrated on 2 Datasets)} \\
\cmidrule(lr){2-4} \cmidrule(lr){5-8}
 & CORE & UCF & Effort & \makecell{UniFace \\ + BleFace} & \makecell{MobSwap \\ + e4s} & \makecell{FaceDan \\ + FSGAN} & \makecell{InSwap \\ + SimSwap} \\
\midrule
UniFace & 0.871 & 0.831 & 0.962 & 0.981 & 0.974 & 0.982 & 0.977 \\
BleFace & 0.844 & 0.827 & 0.873 & 0.941 & 0.925 & 0.939 & 0.928 \\
MobSwap & 0.959 & 0.950 & 0.953 & 0.971 & 0.964 & 0.985 & 0.969 \\
e4s     & 0.680 & 0.732 & 0.982 & 0.988 & 0.982 & 0.989 & 0.983 \\
FaceDan & 0.774 & 0.862 & 0.920 & 0.971 & 0.958 & 0.965 & 0.966 \\
FSGAN   & 0.958 & 0.937 & 0.957 & 0.980 & 0.973 & 0.985 & 0.977 \\
InSwap  & 0.855 & 0.809 & 0.936 & 0.946 & 0.932 & 0.953 & 0.948 \\
SimSwap & 0.724 & 0.647 & 0.926 & 0.948 & 0.938 & 0.950 & 0.954 \\
\midrule
Average & 0.833 & 0.824 & 0.939 & \cellcolor[HTML]{E8F0FE}0.966 & \cellcolor[HTML]{E8F0FE}0.956 & \cellcolor[HTML]{E8F0FE}0.969 & \cellcolor[HTML]{E8F0FE}0.963 \\
\bottomrule
\end{tabular}
\end{adjustbox}
\par\medskip

{(b) Face Reenactment (FR) Category}\par
\begin{adjustbox}{max width=\linewidth}
\begin{tabular}{l | c c c | c c c c c c}
\toprule
\multirow{2}{*}{Testing Dataset} & \multicolumn{3}{c|}{Baseline} & \multicolumn{6}{c}{Ours (Calibrated on 2 Datasets)} \\
\cmidrule(lr){2-4} \cmidrule(lr){5-10}
 & CORE & UCF & Effort & \makecell{FOMM \\ + FV2V} & \makecell{Wav2Lip \\ + MRAA} & \makecell{OneShot \\ + PiRender} & \makecell{TPSMM \\ + LIA} & \makecell{DaGan \\ + SadTalker} & \makecell{MCNet \\ + HyperReenact} \\
\midrule
FOMM         & 0.932 & 0.889 & 0.901 & 0.949 & 0.945 & 0.962 & 0.959 & 0.958 & 0.963 \\
FV2V         & 0.822 & 0.794 & 0.810 & 0.873 & 0.852 & 0.866 & 0.855 & 0.899 & 0.914 \\
Wav2Lip      & 0.899 & 0.874 & 0.733 & 0.861 & 0.881 & 0.885 & 0.896 & 0.882 & 0.883 \\
MRAA         & 0.983 & 0.978 & 0.946 & 0.991 & 0.986 & 0.994 & 0.989 & 0.993 & 0.993 \\
OneShot      & 0.861 & 0.818 & 0.802 & 0.890 & 0.898 & 0.917 & 0.905 & 0.916 & 0.913 \\
PiRender     & 0.901 & 0.862 & 0.804 & 0.943 & 0.916 & 0.924 & 0.918 & 0.935 & 0.920 \\
TPSMM         & 0.816 & 0.792 & 0.758 & 0.860 & 0.822 & 0.865 & 0.849 & 0.869 & 0.888 \\
LIA          & 0.926 & 0.913 & 0.915 & 0.957 & 0.947 & 0.968 & 0.960 & 0.974 & 0.969 \\
DaGan        & 0.731 & 0.753 & 0.742 & 0.821 & 0.789 & 0.834 & 0.812 & 0.830 & 0.854 \\
SadTalker    & 0.715 & 0.662 & 0.594 & 0.710 & 0.666 & 0.701 & 0.687 & 0.681 & 0.672 \\
MCNet        & 0.833 & 0.813 & 0.769 & 0.864 & 0.833 & 0.876 & 0.867 & 0.883 & 0.899 \\
HyperReenact & 0.878 & 0.747 & 0.926 & 0.902 & 0.913 & 0.911 & 0.931 & 0.928 & 0.922 \\
\midrule
Average      & 0.858 & 0.825 & 0.808 & \cellcolor[HTML]{E8F0FE}0.885 & \cellcolor[HTML]{E8F0FE}0.871 & \cellcolor[HTML]{E8F0FE}0.892 & \cellcolor[HTML]{E8F0FE}0.886 & \cellcolor[HTML]{E8F0FE}0.896 & \cellcolor[HTML]{E8F0FE}0.899 \\
\bottomrule
\end{tabular}
\end{adjustbox}
\end{table}

\subsection{Real-world perturbations} We evaluated the robustness of COMPASS under severe common unseen corruptions in \cref{table:real_world_perturbations}: JPEG compression (Quality=80), Gaussian Blur ($\sigma=0.5, k=5$), Gaussian Noise ($\sigma=0.01$), and Resizing ($0.75\times$). In the table below, COMPASS demonstrates remarkable resilience and consistently outperforms the baseline (EFFORT) across all distortions.
\label{real_world_perturbations}

\begin{table}[h]
\centering
\begin{tabular}{lcccccc}
\hline
\rowcolor{lightgray!30} \textbf{Method} & \textbf{Clean} & \textbf{JPEG} (80) & \textbf{Blur} ($
\sigma$=0.5) & \textbf{Noise} (0.01) & \textbf{Resize} (0.75$\times$) & \textbf{Avg.} \\ \hline
EFFORT & 0.940 & 0.927 & 0.936 & 0.913 & 0.920 & 0.927 \\
COMPASS & \textbf{0.966} & \textbf{0.947} & \textbf{0.949} & \textbf{0.930} & \textbf{0.946} &
\textbf{0.948} \\ \hline
\end{tabular}
\caption{Robustness evaluation under real-world perturbations.}
\label{table:real_world_perturbations}
\end{table}

\end{document}